\documentclass[letterpaper]{article} 
\usepackage{aaai23}  
\usepackage{times}  
\usepackage{helvet}  
\usepackage{courier}  
\usepackage[hyphens]{url}  
\usepackage{graphicx} 
\usepackage{natbib}  
\usepackage{caption} 
\usepackage{algorithm}
\usepackage{algorithmic}

\usepackage{amsmath}        
\usepackage{multirow}   
\usepackage{booktabs} 
\usepackage{color}
\usepackage{amssymb}
\usepackage{newfloat}
\usepackage{listings}
\DeclareCaptionStyle{ruled}{labelfont=normalfont,labelsep=colon,strut=off} 
\floatstyle{ruled}
\newfloat{listing}{tb}{lst}{}
\floatname{listing}{Listing}
\title{Target-Aware Tracking with Long-term Context Attention}
\author{
    Kaijie He\textsuperscript{\rm 1}\\
    Canlong Zhang\textsuperscript{\rm 1,\rm 2}\thanks{Corresponding author},
    Sheng Xie\textsuperscript{\rm 1},
    Zhixin Li\textsuperscript{\rm 1,\rm 2},
    Zhiwen Wang\textsuperscript{\rm 3}
}
\affiliations{
    \textsuperscript{\rm 1}School of Computer Science and Engineering, Guangxi Normal University, China\\
    \textsuperscript{\rm 2}Guangxi Key Lab of Multi-source Information Mining and Security, China\\  
    \textsuperscript{\rm 3}School of Computer Science and Technology,  Guangxi University of Science and Technology, China\\    

    hekaijie123@outlook.com, clzhang@gxnu.edu.cn
}

\usepackage{bibentry}

\begin{document}

\maketitle

\begin{abstract}
    Most deep trackers still follow the guidance of the siamese paradigms and use a template that contains only the target without any contextual information, which makes it difficult for the tracker to cope with large appearance changes, rapid target movement, and attraction from similar objects. To alleviate the above problem, we propose a long-term context attention (LCA) module that can perform extensive information fusion on the target and its context from long-term frames, and calculate the target correlation while enhancing target features. The complete contextual information contains the location of the target as well as the state around the target. LCA uses the target state from the previous frame to exclude the interference of similar objects and complex backgrounds, thus accurately locating the target and enabling the tracker to obtain higher robustness and regression accuracy. By embedding the LCA module in Transformer, we build a powerful online tracker with a target-aware backbone, termed as TATrack. In addition, we propose a dynamic online update algorithm based on the classification confidence of historical information without additional calculation burden. Our tracker achieves state-of-the-art performance on multiple benchmarks, with 71.1\% AUC, 89.3\% NP, and 73.0\% AO on LaSOT, TrackingNet, and GOT-10k. The code and trained models are available on https://github.com/hekaijie123/TATrack.
\end{abstract}

\section{Introduction}

Visual target tracking is a fundamental computer vision task. Given an initial position of any target, the tracker is required to evaluate the target state in subsequent each frame of a video. Tracking task faces significant challenges such as the variable appearance, fast movement, attraction from similar objects, etc. Siamese structure based trackers have achieved considerable success, which realizes tracking by using twin networks to represent the target and search image and calculating their similarity. Although the existing trackers have become more and more complex, most of them still originate from Siamese paradigms.

After carefully investigating the existing Siamese trackers, we found that they have more or less inherited certain simple operation from SiamFC \cite{simafc}, and have some drawback as follows: (1) Due to insufficient appearance information, using target template without any background to correlate with the search region will be diffcult to distinguish the real target from the background attractors similar to the target, and also has difficulty in coping with the severe appearance changes; (2) In tracking task, the correlation operation is performed after the backbone network has completely extracted the image features, but the backbone network is originally designed for classification task, so the applicability of the feature extraction to the tracking task is limited to a certain extent. (3) The tracker uses only the optimal model obtained by offline training to predict the target, and the model only knows the target appearance of the initial frame without online updated information. Such a tracker is static and it does not have any perception of the changes that occur in the target state throughout the video sequence. The static tracker lacks perception of the continuous changes in the target and loses robustness in long time tracking. Above drawbacks are not very prominent in simple tracking scenarios, but they will be rapidly enlarged in complex tracking scenarios, so it is necessary to overcome them.
\begin{figure}[t]
\includegraphics[width=0.95\columnwidth]{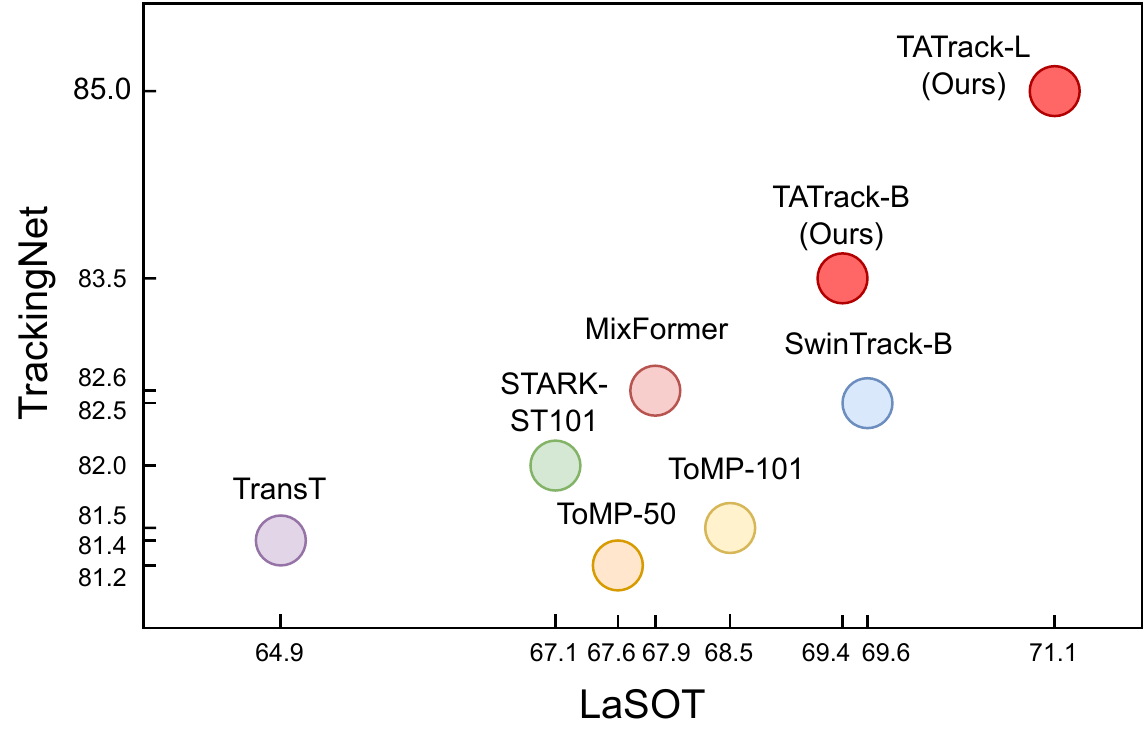} 
\caption{Comparison with other advanced trackers on TrackingNet and LaSOT benchmarks.}
\label{fig0}
\end{figure}
Inspired by above three problems, we propose a long-term context attention mechanism that can simultaneously accept a target template, a historical frame and a current search frame as input in an adaptive weighted fusion way. We embed an improved location encoder in the LCA, which enables the target template, the historical frame and the current search frame to perform self-attention calculation while perform cross-attention calculation with each other. The LCA extensively fuses target and background features of images spanning different times and can effectively extract the location information of the target and the state information around the target. Since the LCA module has correlation calculation and feature extraction functions, we alternately stack multiple LCA and SWA \cite{swin} modules to construct the backbone network suitable for tracking task. With the deepening of layers of the backbone network, the target-aware ability of LCA module will be stronger, that is, the real target will be highlighted while the features of other interfering objects will be weakened. The previous templates are filled with high quality historical frames from the inference process, so we need a reliable quality determination method to select the historical frames. Different from existing online update trackers that use a two-stage inference update approach \cite{tomp} or a two-stage training network structure \cite{mixformer,stark}, we propose a very concise and efficient algorithm that determines whether to update the template according to the classification confidence scores of historical frames, thus achieving high robustness and avoiding large calculational cost like aforementioned two-stage methods. TATracker achieves state-of-the-art performance, shown in Fig. \ref{fig0}.

In summary, our main contributions as follows:
\begin{itemize}
    \item We propose a new cross-frame attention module suitable for fusion interaction of target and its context.
    \item Based on LCA, we build a powerful tracker that has a backbone better suited to the tracking task.
    \item We propose a concise and efficient online updating approach based on classification confidence to select high-quality templates with very low computation burden.
    \item We evaluate our tracker through comprehensive ablation and comparison experiments, and the experimental results verify its effectiveness and advancement.
\end{itemize}

\section{Related Work}

\subsubsection{Tracker Backbone.}
Deep trackers rely heavily on offline training, and more powerful feature extraction networks can capture deeper semantic information about the target. This feature allows twin network architectures to easily gain more powerful performance from each backbone network upgrade. From the early days of siamFC \cite{simafc}, SiamRPN \cite{siamrpn} used AlexNet, then SiamRPN++ \cite{siamrpn++} pioneered the use of the more mature backbone network Resnet, to recent years when Transformer backbone networks started to be used in trackers \cite{swintrack}. In these previous works, the backbone networks used by the tracker were derived from the upstream image classification task, and the direct use of the feature extraction network for the classification task is inefficient for the tracker. The backbone network for the classification task is used to determine the overall category of the image and has no perception of the target and background in the tracking task, which is contrary to the requirement of distinguishing interferers in the tracking task. We propose a target-aware backbone that focuses on the extraction of target features. In addition, we also add auxiliary positioning information as \cite{ZHANG201832} fuses multi-feature information.

\subsubsection{Online Update.}
In the Siamese paradigm, the tracker uses the first frame as a template and remains unchanged, and the performance of the tracker depends entirely on the ability to match the appearance of the target. However, the appearance of the target tends to change continuously over time, and models that are not updated have significant bottlenecks. Guided by this, much work has been done to experiment with online updates. The large network structure of the tracker requires long time and large amount of data for training, and the video history information obtained online alone can hardly be used to accurately update the model parameters of the subject. the ATOM \cite{atom} model is designed with mini-localization branches, and the localization branches are trained at each inference. However, mini-branches have significant performance limits and increase inference time, and have not become mainstream. UpdateNet \cite{updatenet} uses a CNN to add the target template and the cumulative template to the current frame template with certain weights, resulting in a template with continuously changing information, but the template will accumulate contamination over time leading to failure. STMtrack \cite{stmtrack} uses fused information from multiple templates, and updates are taken from the historical template at medium intervals. MixFormer \cite{mixformer,stark}et al. take the training quality branch to score the history frames, which has the disadvantage of secondary training and requires restarting the training quality branch after the training of the main body of the network. ToMP \cite{tomp} uses two templates and takes the most recent frame that meets the conditions as the template, but the inference phase requires two repetitions to ensure performance. While each of these online update approaches possesses relatively obvious limitations, our approach uses historical information about the classification confidence generated during the tracker inference to achieve a way to update templates online without any additional cost.

\section{Method}

\begin{figure*}[t]
\centering
\includegraphics[width=0.9\textwidth]{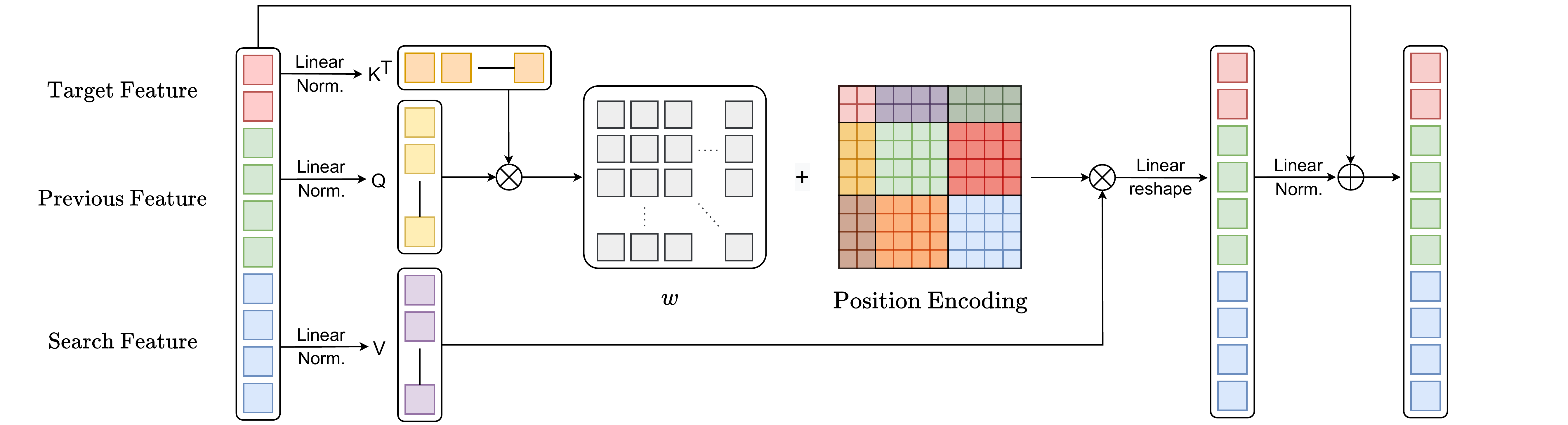} 
\caption{Long-term contextual attention module (LCA) is an efficient multi-image attention operation. It can simultaneously perform feature extracting for each image through self-attention and target searching through cross-attention among images. LCA uses inter-image independent location encoding to divide the attention weight map into TtoT, TtoP, TtoS, PtoT, PtoP, PtoS, StoT, StoP, and StoS from top to bottom and left to right, where T, P, and S represent the target template, the previous template, and the searched image, respectively.}
\label{fig1}
\end{figure*}

\subsection{Long-term Contextual Attention (LCA)}
In this section, we first introduce the proposed Long-term Contextual Attention (LCA) module, which is designed for integrating the information of target and context from multi-frames. Fig. \ref{fig1} shows the overview of the LCA module.

LCA is a powerful attention computation module, which can integrate the features from the target template, previous template, and searched image, perceive the real target from the previous template and search image based on the features of the target template, and reinforce features of the target while weakening interference information. At the same time, the LCA can implicitly find the changes in the search image based on the target state including the appearance and relative position in the previous template, to further exclude the interfering objects similar to the target, thus more accurately capturing the current target state.

In this module, position encoding plays an important role. We know that in the self-attention formula Eq. \ref{eq1}, the self-attention formula without position encoding is Eq. \ref{eq2} and the self-attention formula with absolute position encoding is Eq. \ref{eq3}. $x\in \mathbb{R}^{L\times d}$ is the input feature, p is the absolute position encoding, and $W\in \mathbb{R}^{d\times d}$ is the linear transformation matrix. Q, K and V represent the query, the key and value three mapping matrices, which have the same dimensionality. We did not just directly stitch together the target template, the previous template, and the search image as a whole to calculate the self-attention. Because the relationship between the target template, previous template, and search image will not be distinguished in the formula, the three features will be treated as one big image and the model's ability to construct connections between the three will be limited. Therefore, we must make improvements to the location coding.
\begin{align}\label{eq1}  \text { Attention } &=\operatorname{Softmax}(w)xW^{V}    \end{align} 
\begin{equation}\label{eq2} \text{where} \ \ w=\frac{1}{\sqrt{d}}(x W^{Q}) (x W^{K})^{T} \end{equation}
\begin{align}\label{eq3}  w^{A b s} &=\frac{\left(\left(x+p\right) W^{Q}\right)\left(\left(x+p\right) W^{K}\right)^{T}}{\sqrt{d}}\end{align}
TUPE \cite{tupe} proposes untied absolute positional encoding \ref{eq4} as an alternative to the traditional positional encoding formulation, where $U$ is a linear transformation matrix of learnable absolute positional encoding with dimensionality equal to $W$. This formulation decouples the feature $x$ from the absolute positional encoding $p$. $p$ extracts position information with a learnable independent transformation matrix, unlocking the potential of absolute position coding.
\begin{equation} \label{eq4} \begin{aligned} w^{A b s}&=\frac{1}{\sqrt{2 d}}\left(x W^{Q}\right)\left(x W^{K}\right)^{T} \\&+\frac{1}{\sqrt{2 d}}\left(p U^{Q}\right)\left(p U^{K}\right)^{T} \end{aligned} \end{equation}
We adopt the relative position encoding $r\in \mathbb{R}^{L\times L}$ form used in SWA \cite{swin} Eq. \ref{eq5}, where the $\mathbf{P}_{i-j}$ is the learnable variable de-indexed according to $i-j$.
\begin{equation} \label{eq5} \mathbf{r}_{i j}=\mathbf{P}_{i-j} \end{equation}
We extend the positional encoding designed for a single image to the multi-image case. We first expand the input target template, the previous template, and the features of the search region into a dimension L along the W and H dimension, and concatenate the three together along the L dimension $x=Concat\left(\mathrm{z},\mathrm{pre},\mathrm{x}\right)$, and do the same for the absolute position encoding $p=Concat\left(p_z,p_{pre},p_x\right)$. Divide the relative position encoding $r$ into $n\times m\ =\ 9$ independent regions to compute $ \mathbf{P}_{\ i-j}$, as shown by the position encoding in Fig. \ref{fig1}. Finally, we obtain the LCA attention formula as follows Eq. \ref{eq6}.
\begin{align} \label{eq6} \begin{gathered}  \text { LCA } =\operatorname{Softmax}(w+a+r)xW^{V}   \\
w =\frac{1}{\sqrt{2d}}(x W^{Q}) (x W^{K})^{T} \\  
a = \frac{1}{\sqrt{2 d}}\left(p U^{Q}\right)\left(p U^{K}\right)^{T}  \\
r_{nm,i j} =\mathbf{P}_{nm,i-j}  \end{gathered} \end{align} 

\subsubsection{Discussions:}
Why do we introduce additional prior templates? (1) We introduce rich background information through the previous templates to locate the target with the help of the relative position of the previous target in the background. (2) The addition of prior templates enables online updates, and prior templates contain the state of the target at more recent time points for more accurate regression of the target. Why did we not add the background to the target template? Because, the previous template is temporally closer to the current frame, and the initial target is not important to include the background in the target template, and including only the non-updated targets can emphasize the consistency of the tracker's tracking of the target.

\subsection{Overall}

\begin{figure*}[t]
\centering
\includegraphics[width=0.9\textwidth]{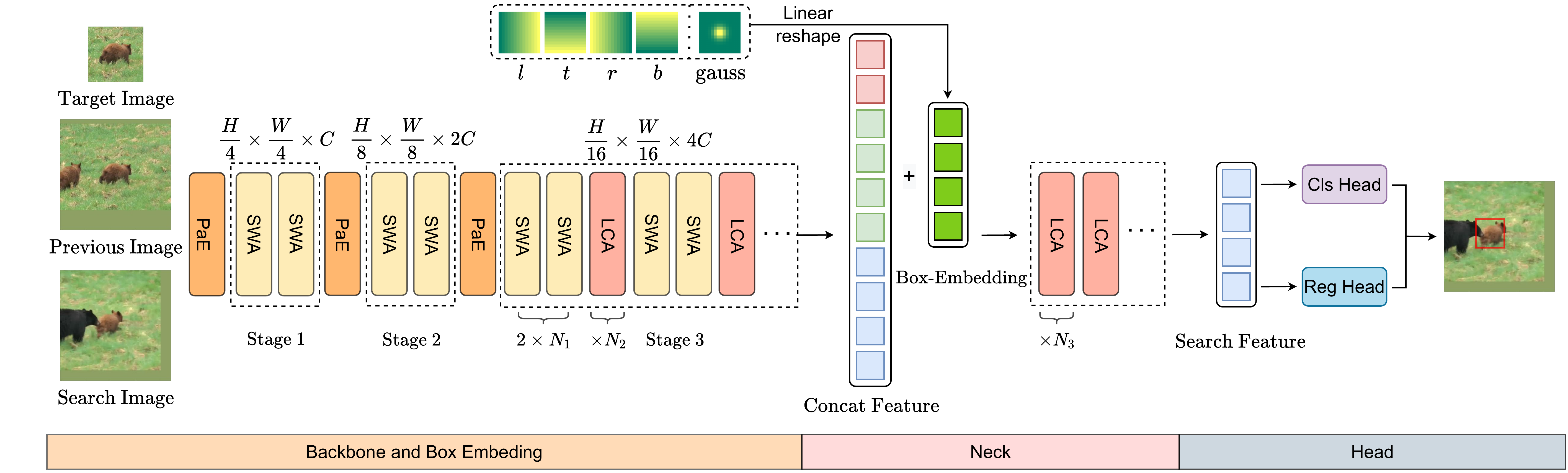} 
\caption{We construct a target-aware backbone network based on alternating stacks of LCA and shift window attention (SWA) module of Swin-Transformer. The target-aware features extracted by the backbone network are further refined by a neck consisting of multiple layers of LCAs to refine the state information of the target. Finally, only the features of the search image part are taken for the feature maps of the regression and classification head.}
\label{fig2}
\end{figure*}

The overall structure of the model is shown in Fig. \ref{fig2}.

\subsubsection{Backbone.}
The backbone network accepts the target template $z \in \mathbb{R}^{H_z\times W_z\times 3}$, the previous template $pre\in \mathbb{R}^{H_{pre}\times W_{pre}\times 3}$ and the search image $x\in \mathbb{R}^{H_x\times W_x\times 3}$ of the input. To better match the LCA, we choose the SWA with the same transformer structure to build the feature extraction network. In the first stage of the feature extraction network, we patch and embed the image features into $\frac{H}{4}\times\frac{W}{4}\times C$ feature tokens using a convolution kernel of size 4. The token expands both $H$, $W$ dimensions into length $L$ in the transformer operation. and then passed through two SWA modules. PaE is used before each subsequent stage, and $H$, $W$ is halved while $C$ is doubled. In the stage3, we use two SWA modules in a group with LCA modules stacked alternately. We use two SWA modules as a group because two SWA modules complete a window shift and recovery. The target features, previous features and search features are individually passed through PaE and SWA modules in turn, and the three are input into LCA after stitching along the $L$ dimension, and the output of LCA is reduced to three tokens in turn by SWA to extract features. Finally, the sequence of $\frac{H}{16}\times\frac{W}{16}\times4C$ tokens is obtained from the backbone network.

\subsubsection{Neck.}
In this stage, we add the encoded information about the target location and target size of the previous template. We increase the robustness in training by randomly dithering regions of the previous template, where the target is not always in the center of the image. The previous template contains larger regions other than the target, and the inclusion of Gaussian localization information is necessary to avoid ambiguities induced by similar objects. Also, we adopt a similar approach by adding the ltrb representation \cite{tomp} of the length and width information to help the model prediction. For the bounding box of the previous frame $b_{pre}=\left\{b^{x_1},b^{y_1},b^{x_2},b^{y_2}\right\}$, we first denote each position of the feature map by $\left(k^x,k^y\right)$ . Then we get the formula Eq. \ref{eq7} for the bounding box information $d_{pre}=\left(l,t,r,b\right)$ represented by the four sides.
\begin{align} \label{eq7} \begin{array}{ll}
    l=k^{x}-b^{x_1}/s  , & r=b^{x_2}/s-k^{x} , \\ 
    t=k^{y}-b^{y_1}/s  , & b=b^{y_2}/s-k^{y} ,
\end{array} \end{align} 
where $s=16$ is the multiple of image mapping to feature map reduction. $d_{pre}\in \mathbb{R}^{H \times W \times 4}$, $\psi$ is the multilayer perceptron that maps the dimension of d from 4 to C to get the size information embedding of the target. The Gaussian position map $y_{pre}\in \mathbb{R}^{H\times W\times 1}$ mentioned earlier is multiplied by the learnable weight $w\in \mathbb{R}^{1 \times 1 \times C}$ using the broadcast mechanism to get the position information embedding of the previous target. Box-embedding is given by the target size and the target location embedding are directly summed by the following equation:
\begin{align} \label{eq8} \text{Box-embedding}  = w \cdot y_{pre} + \operatorname{MLP}(d_{pre}) \end{align} 
Then we successively overlap with multiple LCA modules to further process the information of the target. In the last layer of LCA, we keep only the attention computation of the search features to the target template and the previous template, then pass the search feature map to the classification head and regression head.

\subsubsection{Head and Loss.}
The head network contains two branches: the bounding box regression head and the classification head. The regression map $\mathbb{R}^{H\times W\times 4}$,and the corresponding map $\mathbb{R}^{H\times W\times 1}$ for the prediction are obtained from the search feature map through the three-layer perceptron, respectively. The regression loss uses the common generalized IoU loss \cite{giou}, and the classification loss we use varifocal loss \cite{varifocal}, which uses the currently more popular IoU-aware design. the core idea of IoU-aware is to replace the IoU score with the positive label value of the classification, the traditional positive sample value of the classification is labeled as 1, the improvement relates the regression task to the classification task. varifocal loss is formulated as follows:
\begin{align} \label{eq9} \operatorname{VFL}(\mathrm{p}, \mathrm{q})= \begin{cases}-q(q \log (p)+(1-q) \log (1-p)) & q>0 \\ -\alpha p^{\gamma} \log (1-p) & q=0\end{cases} \end{align} 
$p$ is the classification prediction while $q$ is the IoU score of the target. Our total loss function is as follows:
\begin{align} \label{eq10} \mathcal{L} =\lambda_{\text {cls }} \mathcal{L}_{\mathrm{cls}}\big(p,\operatorname{IoU}(\hat{b}, b)\big)+\lambda_{\text {giou }} \mathcal{L}_{\text {giou }}(\hat{b}, b) \end{align} 
For the weights $\lambda_{\mathrm{cls}}$ is set to 1.5 and $\lambda_{\mathrm{giou}}$ is set to 1.5.
\subsection{Online Update Strategy}

A common way to pick an update template is to retrain a quality judgment branch, but such an approach requires training the model twice. And almost all update approaches set a fixed hyperparameter as a threshold, and update only when the quality confidence is higher than this static threshold. We believe that updating the template should follow two principles, (1) The updated template should be as close to the current frame as possible in time to ensure that the updated template has the most similar state to the current frame. (2) The updated template should be as high quality as possible, with good recognition and accuracy. In this regard, we propose a dynamic thresholding algorithm with a simple classification confidence level to select the ideal historical template. The first algorithm uses the classification confidence historical average as the threshold value, Eq. \ref{eq11} as follows:
\begin{align} \label{eq11} mean = \sum_{i}^{n}s_i/n \end{align} 
$n$ is the current frame serial number n and $s_i$ is the classification confidence of the ith frame. The mean value as a threshold has a low update criterion and may be selected as close as possible to the historical template of the current frame. This approach shows relatively reliable performance in the got-10k and TrackingNet datasets, but underperforms in the long sequence tracking dataset Lasot. We analyzed the reasons for this; as the target deformation and the environmental changes it faces tend to get more and more complex with increasing time, the model keeps accumulating errors during the tracking process, and even loses the target forever after losing it midway. This is not obvious on got-10k and TrackingNet short series datasets, but long series benchmarks like LaSOT are more likely to encounter prolonged occlusion or target disappearance, and using only lower thresholds will tend to update to the wrong template. Wrong updates will continuously reduce the classification confidence in subsequent tracking, making the threshold of the mean formula invalid. Therefore, we further propose an improved calculation method by proposing a threshold formula with penalty Eq. \ref{eq12}.

The results of the threshold formula with penalty are more dependent on the prior classification confidence scores, and the results are more stable compared to the mean formula. Keeping the threshold higher allows the model to resist erroneous template updates when encountering long periods of target disappearance and occlusion.
\begin{equation}\label{eq12} \begin{aligned} p\_mean &=  \sum_{m}^n \big(\sum_{i}^{m}s_i/m \big) /n \\ &= \big[\big(s_1 + \frac{s_1}{2} + \dots \frac{s_1}{n}\big) \\ &+ \big(\frac{s_2}{2} + \dots \frac{s_2}{n}\big)+\dots \big(\frac{s_n}{n}\big) \big]/n\end{aligned}\end{equation}
The results of the threshold formula with penalty are more dependent on the prior classification confidence scores, and the results are more stable compared to the mean formula. Keeping the threshold higher allows the model to resist erroneous template updates when encountering long periods of target disappearance and occlusion.

\section{Experiments}

\subsection{Implementation details}

\begin{table}[t]
\small      
\centering
\resizebox{.95\columnwidth}{!}{      
\renewcommand{\arraystretch}{1.5}       
\begin{tabular}{c|c|c|c}
\hline

    & TATrack-S & TATrack-B & TATrack-L \\
\hline
Target Image & $112 \times 112$ & $112 \times 112$ & $192 \times 192$ \\
\hline
Previous Image & $224 \times 224$ & $224 \times 224$ & $384 \times 384$ \\
\hline
Search Image & $224 \times 224$ & $224 \times 224$ & $384 \times 384$ \\
\hline
Backbone & $\begin{bmatrix}N_1=3\\ N_2=2\\C=96\end{bmatrix}$ & $\begin{bmatrix}N_1=9\\ N_2=8\\C=128\end{bmatrix}$ & $\begin{bmatrix}N_1=9\\ N_2=8\\C=128\end{bmatrix}$ \\
\hline
Neck & $N_3=4$ & $N_3=8$ & $N_3=8$ \\
\hline
MACs & 13.1 G& 45.1 G& 162.4 G\\
\hline
Param & 24.5 M & 112.8 M & 112.8 M\\
\hline
Speed(V100) & 29.6 FPS & 14.1 FPS & 6.6 FPS \\
\hline
    
\end{tabular}}
\caption{The network structure parameters of TATrack-S, TATrack-B, and TATrack-L. The number of $N_1$, $N_2$, $N_3$, $C$ corresponds to Fig. \ref{fig2}}
\label{table1}
\end{table}
Our tracker was implemented on Python 3.9 and pytorch 1.11.0, trained on 2 Tesla A100 GPUs. The different sizes of TATrack are shown in Tab. \ref{table1}, on PaE and SWA modules, TATrack-S, TATrack-B, and TATrack-L are loaded with pre-training weights of Swin-Tiny, Swin-Base, and Swin-Base384, respectively. We used TrackingNet, LaSOT, COCO and GOT-10k multiple training sets for joint training.

\subsection{Comparison with the State-of-the-art Trackers}

\begin{table*}[t]
\small      
\centering
\resizebox{0.95\textwidth }{!}{      
\renewcommand{\arraystretch}{1.5}       
\begin{tabular}{c|c|ccc|ccc|ccc}
\hline

{ \multirow{2}{*}{Method} }& { \multirow{2}{*}{Published} }&\multicolumn{3}{c|}{GOT-10k} &\multicolumn{3}{c|}{TrackingNet} &\multicolumn{3}{c}{LaSOT}  \\       
\cline{3-11}
&& AO(\%)	& $SR_{50}$(\%)	& $SR_{75}$(\%)	& AUC(\%)	& $P_{norm}$(\%)	& P(\%)	& AUC(\%)	& $P_{norm}$(\%)	& P(\%)	 \\   
\hline
TATrack-L	&Ours&\underline{79.2}	&\underline{88.6}	&\underline{78.3}	&\textcolor{red}{85.0}	&\textcolor{red}{89.3}	&\textcolor{red}{84.5}	&\textcolor{red}{71.1}	&\textcolor{red}{79.1}	&\textcolor{red}{76.1} \\
TATrack-B	&Ours&\underline{77.3}	&\underline{87.8}	&\underline{74.1}	&\textcolor{blue}{83.5}	&\textcolor{blue}{88.3}	&\textcolor{blue}{81.8}	&69.4	&78.2	&\textcolor{blue}{74.1}   \\
TATrack-S	&Ours&\underline{74.3}	&\underline{84.5}	&\underline{70.6}	&81.8	&86.9	&79.7	&68.1	&77.2	&72.2   \\
TATrack-B*	&Ours&\textcolor{red}{73.0}	&\textcolor{red}{83.3}	&\textcolor{red}{68.5}  &-&-&-&-&-&-    \\
\hline
MixFormer\cite{mixformer}   &CVPR22   &\textcolor{blue}{71.2}&79.9&\textcolor{blue}{65.8}    &82.6&87.7&81.2  &67.9&77.3&73.9 \\
ToMP\cite{tomp}    &CVPR22   &-&-&-  &81.2&86.2&78.6 &67.6&78.0&72.2 \\
SBT-B\cite{sbt}   &CVPR22   &69.9&\textcolor{blue}{80.4}&63.6 &-&-&-  &65.9&-&70.0    \\
SwinTrack-B\cite{swintrack}   &arXiv21   &69.4&78.0&64.3   &82.5&87.0&80.4 &\textcolor{blue}{69.6} &\textcolor{blue}{78.6} &\textcolor{blue}{74.1} \\
KeepTrack\cite{keeptrack}   &ICCV21   &-&-&-  &-&-&-  &67.1&77.0&70.2 \\
STARK\cite{stark}   &ICCV21   &68.8&78.1&64.1  &82.0&86.9&-    &67.1&77.0&-    \\
TransT\cite{trant}  &CVPR21   &67.1&76.8&60.9  &81.4&86.7&80.3 &64.9&73.8&69.0 \\
Ocean\cite{ocean}   &ECCV20   &61.1&72.1&47.3 &-&-&-  &56.0&65.1&56.6 \\
SiamPRN++\cite{siamrpn++}   &CVPR19   &51.7&61.6&32.5 &73.3&80.0&69.4 &49.6&56.9&49.1 \\
SiamFC\cite{simafc}  &ECCVW16  &34.8&35.3&9.8  &57.1&66.3&53.3 &33.6&42.0&33.9     \\
\hline

    
\end{tabular} }
\caption{Comparison with the state of the art on the GOT-10k,TrackingNet and LaSOT. The underlined results in GOT-10k are not involved in the comparison because the models are trained based on multiple datasets. TATrack-B* is trained on the GOT-10k training set only.}
\label{table2}
\end{table*}

\subsubsection{GOT-10k.}
GOT10k \cite{got} is a large benchmark containing 560 classes of motion objects and 87 classes of motion patterns, and places more emphasis on the regression accuracy of the tracker on the target.GOT-10k officially requires that the tracker be trained based only on the training set of GOT-10k, and we followed this guidance.GOT-10k provides 180 test sequences with an average sequence length of 150, and officially does not disclose the true annotation of the test sequences, and we obtained the tracking metrics by submitting the raw tracking results for online evaluation Tab. \ref{table2}. 

\subsubsection{TrackingNet.}
TrackingNet \cite{trackingnet} contains 30312 video sequences, videos captured from real-life filmed YouTube content. trackingNet provides 511 test videos with an average sequence length of 441 frames, and no real annotation of the test sequences is publicly available. We submit the raw data to an official online evaluation service, and we achieve state-of-the-art performance in TATrack-L Tab. \ref{table2}.

\subsubsection{LaSOT.}
LaSOT \cite{lasot} is a large benchmark for long sequences, it contains 280 test sequences averaging 2500 frames, and the challenge of LaSOT is the robustness of long-term tracking. We applied the p-mean algorithm on LaSOT to calculate the threshold values and achieved state-of-the-art performance Tab. \ref{table2}.



\subsection{Ablation Study and Analysis}
We did ablation experiments on the components of TATrack and we analyzed the contribution of each separable component in the model. We designed different combinations of templates to verify that templates with background are necessary for the tracker, and we demonstrated the effectiveness of the dynamic thresholding algorithm by comparing multiple update methods. Both the ablation experiments and the comparison experiments were performed under the TATrack-S model.

\begin{figure*}[t]
\centering
\includegraphics[width=0.95\textwidth]{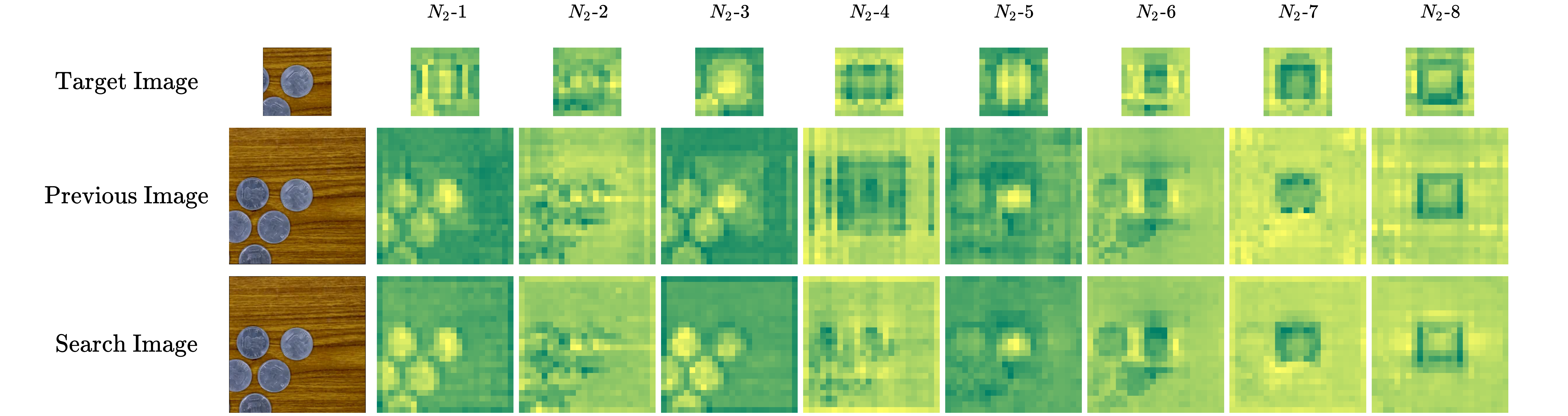} 
\caption{Visualization of the feature maps output from each layer of LCA in the Target-aware backbone network.}
\label{fig3}
\end{figure*}

\begin{table}[t]
\centering
\resizebox{.95\columnwidth}{!}{      
\renewcommand{\arraystretch}{1.5}       

\begin{tabular}{c|c|cc|cc}
\hline
{ \multirow{2}{*}{Modification} }&GOT-10k&\multicolumn{2}{|c|}{TrackingNet} &\multicolumn{2}{c}{LaSOT}  \\
\cline{2-6}
&AO&AUC&P&AUC&P    \\
\hline
TATrack-S   &\textbf{74.3}&81.8&\textbf{79.7}&\textbf{68.1}&\textbf{72.2}  \\
Swin Bac.   &72.9&81.5&79.3&67.1&71.4  \\
No Pos.     &71.3&81.1&78.8&66.7&70.5  \\
No ltrb.    &73.9&\textbf{82.0}&\textbf{79.7}&67.8&72.0  \\
No gauss.   &73.7&81.7&79.2&66.9&70.7  \\
\hline
\end{tabular}}
\caption{Ablation studies on TATrack-S.}
\label{table3}
\end{table}

\subsubsection{LCA Ablation.}
Swin backbone means we remove the LCA module in backbone and use the first three stages of swin transformer to extract features, so that the image loses target perception during the feature extraction. We can see that Tab. \ref{table3} all the metrics in TrackingNet, LaSOT, and GOT-10k are degraded, and we only need to add two layers of LCA in the TATrack-S backbone to get a direct performance improvement. For the third line, we remove all the location codes extended in the LCA module in backbone and neck. The metrics of the three datasets show significant degradation, demonstrating that the absence of location encoding divided independently by image relationships poses significant difficulties for the model to construct associations between multiple images.
\subsubsection{Box-embedding.}
We examined the impact of the box embedding in the previous template Tab. \ref{table3}. When we removed ltrb, there was a small decrease in the tracking metrics of GOT-10k and LaSOT and a small improvement in the AUC of TrackingNet. The ltrb embedding has a more limited improvement on the model, probably due to the fact that LCA already has a sufficiently accurate regression on the target. And there is an error in the prediction of the box during inference, and ltrb may introduce the error of the previous template into the prediction of the current frame. When gauss with localization function is removed, the performance degradation is more significant than removing ltrb. It indicates that the importance of localization information is higher in templates that contain background.

\subsubsection{With or Without Background.}

\begin{table}[t]
\centering
\resizebox{.95\columnwidth}{!}{      
\renewcommand{\arraystretch}{1.5}       

\begin{tabular}{c|c|cc|cc}
\hline
{ \multirow{2}{*}{Modification} }&GOT-10k&\multicolumn{2}{|c|}{TrackingNet} &\multicolumn{2}{c}{LaSOT}  \\
\cline{2-6}
&AO&AUC&P&AUC&P    \\
\hline
TATrack-S   &\textcolor{blue}{74.3}&\textcolor{blue}{81.8}&\textcolor{blue}{79.7}&\textcolor{blue}{68.1}&\textcolor{blue}{72.2}  \\
Only target.   &72.1&80.8&78.0&65.7&68.2  \\
Both background.     &\textcolor{red}{74.7}&\textcolor{red}{82.2}&\textcolor{red}{80.1}&\textcolor{red}{68.5}&\textcolor{red}{72.5}  \\

\hline
\end{tabular}}
\caption{Comparison with the dual-template scheme with only the target and the dual-template scheme with both including the background.}
\label{table4}
\end{table}
In Tab. \ref{table4} we compare the two experimental setups using TATrack. the Only target scheme removes the background from the previous template and does not use box-embedding, which results in a significant drop in performance metrics for all datasets, but has a faster speedup. Because the background is removed and box-embedding information cannot be introduced, this scenario is not a fair comparison. In the scenario where both templates contain background, we add background to the target and the target template does not use box-embedding information. We can see that the template with the background is higher in all performance metrics than the template without the background, even for the target template that is not updated. Considering the performance flatness, we finally chose the compromise between the target-only template and the previous template with the background, which still shows the huge potential of introducing the background in the template for performance improvement.

\subsubsection{Update Experiment.}

\begin{table}[t]
\centering
\resizebox{.95\columnwidth}{!}{      
\renewcommand{\arraystretch}{1.5}       

\begin{tabular}{c|ccc|ccc}
\hline
{ \multirow{2}{*}{Modification} }&\multicolumn{3}{|c|}{GOT-10k} &\multicolumn{3}{c}{LaSOT}  \\
\cline{2-7}
&AO&$SR_{50}$&$SR_{75}$&AUC&$P_{norm}$&P    \\
\hline
No Update   &71.4	&81.3	&66.3	&66.7	&75.8	&70.8  \\
Update the last   &68.3	&77.7	&63.8	&61.0	&68.5	&64.2  \\
Mean.     &\textcolor{red}{74.3}	&\textcolor{red}{84.5}	&\textcolor{red}{70.6}	&66.1	&75.0	&69.8  \\
P-mean. &74.0	&84.3	&70.0	&\textcolor{red}{68.1}	&\textcolor{red}{77.2}	&\textcolor{red}{72.2 }   \\

\hline
\end{tabular}}
\caption{Effects of different update strategies on long and short series datasets.}
\label{table5}
\end{table}

We use a typical short sequence dataset GOT-10k and a long sequence dataset LaSOT to validate our experiments. As shown in Tab. \ref{table5}, (1) no update strategy is adopted and the initial frame is used as the previous template, and this scheme achieves an ordinary performance. (2) is to take a fixed update of the previous frame as the previous template, the performance shows a significant drop, indicating that a low-quality template will be disastrous to the tracker. (3) We adopt the historical confidence mean as the dynamic threshold method to update the template, and we can see that better performance can be obtained in GOT-10k which requires higher regression accuracy. This is because the mean value method as a threshold can achieve a good balance of appropriately skipping low quality templates and keeping the previous templates close to the current frame. However, the threshold of the mean value is too low to resist the long-time target disappearance for long sequence datasets LaSOT. (4) Using the historical mean with penalty will appropriately raise the criteria to resist prolonged low-quality templates and the results are more stable. P-mean shows good performance on LaSOT and has no significant negative impact on short series datasets.

\subsubsection{Visualization of LCA.}
To explore how LCA plays a target-aware role in the backbone network, we visualize the output feature maps of the LCA module in the TATrack-L backbone network. We calculate the average of the target features, previous features, and search features of the LCA output on the C channel and adjust them into a response map of $\mathbb{R}^{H\times W\time 1}$. Through the visualization in Fig. \ref{fig3}, we can conclude that (1) LCA can determine the target location in the previous template and search image layer by layer. (2) LCA can exclude interfering objects layer by layer.

\subsection{Limitations}
The use of multiple templates, especially the previous templates containing background, makes our model run low. TATrack-Small becomes less tiny after using Swin-Tiny pre-training weights. we can see from the experiments that templates containing background have a greater potential to improve the tracker performance. Therefore, in future work we will consider using a transformer module that optimizes the computational effort to better mine templates with background to achieve more powerful performance. Also, we will continue to explore low-cost in new update methods.

\section{Conclusion}
We propose the Long-Term Contextual Attention module, a fusion module for fusing target and background information over multiple time frames. Taking advantage of LCA's ability to simultaneously extract features and compute correlations across images, we propose TATrack with target-awareness. to allow simple and efficient updating of templates, we propose an online update algorithm for dynamic thresholding.

\section{Acknowledgements}
This work is supported by National Natural Science Foundation of China (Nos. 62266009, 61866004, 62276073, 61966004, 61962007), Guangxi Natural Science Foundation (Nos. 2018GXNSFDA281009, 2019GXNSFDA245018, 2018GXNSFDA294001), Guangxi Collaborative Innovation Center of Multi-source Information Integration and Intelligent Processing, and Guangxi "Bagui Scholar" Teams for Innovation and Research Project.

\bibliography{aaai23}

\end{document}